\documentclass[10pt,twocolumn,letterpaper]{article}

\usepackage{cvpr}              

\usepackage{graphicx}
\usepackage{amssymb}
\usepackage{booktabs}
\usepackage{bm}
\usepackage{amsmath,amsfonts,amssymb}
\usepackage{multirow}
\usepackage[accsupp]{axessibility}
\usepackage[switch]{lineno}  %
\usepackage{fancyhdr,graphicx,amsmath,amssymb}
\usepackage{subfloat}
\usepackage[linesnumbered,ruled,vlined]{algorithm2e}
\SetKwInput{KwData}{Input}
\SetKwInput{KwResult}{Output}

\usepackage[pagebackref,breaklinks,colorlinks]{hyperref}
\newcommand{\tran}{\intercal}

\usepackage[capitalize]{cleveref}
\crefname{section}{Sec.}{Secs.}
\Crefname{section}{Section}{Sections}
\Crefname{table}{Table}{Tables}
\crefname{table}{Tab.}{Tabs.}

\def\authorBlock{
    Wei Zhu$^{1,2}~$\thanks{Work was done while Wei Zhu interned at PAII Inc.}~~
    Le Lu$^3$~~
    Jing Xiao$^4$~~ 
    Mei Han$^1$~~
    Jiebo Luo$^2$~~
    Adam P. Harrison$^1$  \\
    $^1$~PAII Inc.~$^2$~University of Rochester~$^3$~Alibaba DAMO Academy~$^4$~PingAn Insurance Group\\
    {\tt\small \{zwvews, tiger.lelu, jiebo.luo, adam.p.harrison\}@gmail.com}\\ 
    {\tt\small xiaojing661@pingan.com.cn, hanmei613@paii-labs.com}
}
\begin{document}

\title{Localized Adversarial Domain Generalization}
\author{\authorBlock}
\maketitle

\begin{abstract}
Deep learning methods can struggle to handle domain shifts not seen in training data, which can cause them to not generalize well to unseen domains. This has led to research attention on domain generalization (DG), which aims to the model's generalization ability to out-of-distribution. Adversarial domain generalization is a popular approach to DG, but conventional approaches (1) struggle to sufficiently align features so that local neighborhoods are mixed across domains; and (2) can suffer from feature space over collapse which can threaten generalization performance. To address these limitations, we propose localized adversarial domain generalization with space compactness maintenance~(LADG) which constitutes two major contributions. First, we propose an adversarial localized classifier as the domain discriminator, along with a principled primary branch. This constructs a min-max game whereby the aim of the featurizer is to produce locally mixed domains. Second, we propose to use a coding-rate loss to alleviate feature space over collapse. We conduct comprehensive experiments on the Wilds DG benchmark to validate our approach, where LADG outperforms leading competitors on most datasets. \url{https://github.com/zwvews/LADG}
\end{abstract}

\section{Introduction}
Deep neural networks can suffer from poor generalization performance on out-of-distribution (OOD) data from unseen domains, \ie{}, from \textit{domain shift}. For this reason, techniques to improve generalization abilities have gained increasing attention over the past decade, \eg{} domain generalization~(DG) and domain adaptation~(DA). DA requires access to data from testing domains during training. In contrast, DG aims to construct a generalized model by exposing the training process to multiple training domains without exposure to OOD data from testing domains. As such, DG can reflect challenges in many real-life applications.
\begin{figure}
\centering
\includegraphics[width=0.45\textwidth]{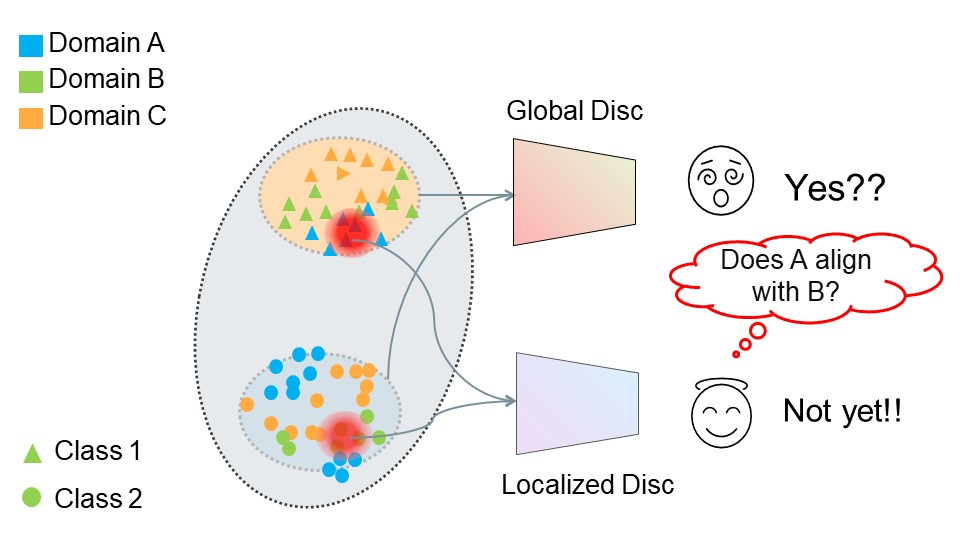}
\caption{Localized discriminator can align two domains in a more fine-grained way.}
\label{fig:brief}
\end{figure}

Different DG methods have been proposed, including empirical risk minimization (ERM)-based methods~\cite{sagawa2019distributionally,zhang2017mixup,zhou2021mixstyle}, meta-learning based methods~\cite{li2018learning}, domain-invariant representation based methods~\cite{ganin2016domain,long2017conditional,sun2017correlation,li2018domain}, invariant risk minimization based methods~\cite{arjovsky2019invariant,zhang2020adaptive}, and gradient agreement methods \cite{fishshi2021gradient,rame2021fishr}. Among them, domain invariant representation based methods, specifically adversarial domain generalization (ADG) methods, seem to the most popular according to recent literature~\cite{sicilia2021domain,gulrajani2020insearch}. ADG methods are inspired by generative adversarial networks~\cite{goodfellow2014generative} and learn a common feature space by adversarial learning for training domains~\cite{ganin2016domain}---the common feature space is expected to help generalization to unseen domains~\cite{ganin2016domain}. Although intuitively reasonable and technically sound, most of these methods show little performance gain over the baseline ERM in practice as indicated by recent benchmarks~\cite{gulrajani2020insearch,koh2021wilds}. 

In this work, we argue that two issues limit the performance impact of current ADG methods. First, we find that ERM feature representations are surprisingly already roughly aligned, with features clustered class-wisely regardless of their domain labels. The discrepancy between domains can be observed at local regions where local neighborhoods are not mixed across domains. Since ADG methods operate under the assumption of significant domain shift, this less obvious domain-level discrepancy challenges conventional ADG methods. Second, we measure the compactness of the feature space using three different metrics and found that the featurizer may trivially fool the discriminator by over collapsing the feature space. The collapsed feature space can cause overfitting~\cite{yu2020learning}. 

To address these limitations, we propose localized adversarial domain generalization with space compactness maintenance~(LADG). As shown in Fig. \ref{fig:brief}, LADG incorporates a localized domain classifier~\cite{bischl2013benchmarking} with adversarial learning. Since the predictions of local classifiers are made according to samples around a target sample, a local domain classifier can thus be used to describe the quality of alignment of local regions. We adopt label propagation in this paper as it is effective and differentiable~\cite{liu2018learning,zhang2019adaptive}. We also outline how to integrate localized classification properly within a generator loss that encourages neighbor hood mixing. To alleviate feature space over collapse, LADG measures and maintains the compactness of the feature space using a differentiable coding ratio~\cite{yu2020learning}. 

We summarize our contributions as follows:
\begin{enumerate}
    \item We argue that ERM can already roughly align training domains with a linear primary task predictor, which undermines the assumption of most existing ADG methods. We also observe that applying ADG to the feature space can lead to feature space over collapse.
    \item We propose localized adversarial domain generalization~(LADG) to alleviate these two limitations. LADG adopts a principled label propagation as the domain discriminator to allow a fine-grained domain alignment and penalizes the space collapse caused by adversarial learning.
    \item We conduct extensive experiments on benchmark datasets to verify our observations and show the effectiveness of the proposed method.
\end{enumerate}

\section{Related Work}
DG is an active research topic with varied approaches. For instance, some methods enhance the generalization ability of ERM directly, \eg{}, using mixup-based methods to generate novel data composed of mixtures of domains~\cite{zhang2017mixup,zhou2021mixstyle,wang2020heterogeneous,wu2020dual}. Other methods try to improve the performance over the worst groups or domains~\cite{sagawa2019distributionally,huang2020self}. Recently, self-supervised learning has also been applied to DG, which is inspired by its success on general representation learning~\cite{kim2021selfreg,motiian2017unified,dou2019domain}. Nonetheless, domain-invariant learning has attracted increasing attention for DG, and can be categorized as either representation invariant~\cite{sun2017correlation,ganin2016domain,long2017conditional,zhang2020adaptive,gong2016domain,li2018domain,shankar2018generalizing}, predictor invariant~\cite{arjovsky2019invariant,krueger2021out,ahmed2020systematic,chang2020invariant}, and gradient invariant methods~\cite{fishshi2021gradient,rame2021fishr,parascandolo2020learning}.

We focus on domain invariant representation learning, specifically adversarial domain generalization (ADG)~\cite{ganin2016domain,long2017conditional,li2018domain,matsuura2020domain,albuquerque2020adversarial,zhou2020deep} in this paper. Domain  adversarial neural network (DANN) adopts a gradient reversal layer and updates the featurizer to fool the domain discriminator by generating domain-invariant representations~\cite{ganin2016domain}. Conditional DANN (CDANN)'s discriminator takes the primary task label into consideration when distinguishing samples from different domains~\cite{long2017conditional,li2018deep}. MMD-AAE adopts maximum mean discrepency (MMD) to align domains, matching the latent representation to a prior distribution by adversarial learning~\cite{li2018domain}. Matsuura \textit{et al.} extend DANN \cite{matsuura2020domain} with unknown domain labels.  Zhou \textit{et al.} adopt a domain transformation network to augment the training data to different training domains for adversarial learning~\cite{zhou2020deep}. Sicilia \textit{et al.} propose to interpolate the feature space by using the gradient of the domain discriminator~\cite{sicilia2021domain}. All these works operate under the assumption of significant domain shifts in the feature space between training domains. However, we empirically observe that the domain-level discrepancy is not as obvious as expected under standard benchmark domain generalization settings. As such, LADG is developed to align the domains in a more fine-grained and local way. Moreover, we propose a loss to prevent space collapse that could potentially benefit all ADG methods and also other fields using adversarial learning, \eg{}, domain adaptation~\cite{zhang2019bridging} and adversarial debiasing~\cite{kim2019learning,zhang2018mitigating}.

\section{Adversarial Domain Generalization}
We focus on the problem of domain generalization. In this setting, the training set, $\mathcal{D}_{tr}=\lbrace x_i, y_i,e_i\rbrace_i^n$, is composed of data from $S$ different domains, \ie{}, $\mathcal{D}_{tr}=\lbrace \mathcal{D}_s \rbrace_s^S$, where $D_s=\lbrace x_i, y_i, s\rbrace_{i=1}^{n_s}$ selects the samples from the $s$-th domain, $x_i$ is the training data, $y_i$ denotes the ground-truth label for the primary task, and the categorical value, $e_i$, denotes the domain label. We note that the primary task label $y_i$ can take different forms for different tasks, \eg a categorical value for classification and a continuous value for regression. We adopt a deep neural network to handle the problem, which contains a featurizer $\phi$ and a predictor $w$. The features extracted by $\phi$ are denoted as $H = \lbrace H_s \rbrace_s^S$, where $H_s = \lbrace \mathbf{h}_{i} \rbrace_{i=1}^{n_s} = \lbrace \phi(x_i) \rbrace_{i=1}^{n_s}$ selects the learned features for $s$-th domain. If the primary task is classification, we can also use $H^y = \lbrace \phi(x_i) ~\forall~y_i=y \rbrace_{i=1}^{n_y}$ to denote the features extracted from samples of the $y$-th class. 

It is reasonable and common to implement the predictor, $\phi$, using a single \textit{linear layer} following standard benchmarks~\cite{gulrajani2020insearch}. We will follow the same practice, which will allow us to directly measure the impact of our contributions compared to alternative approaches. The objective of domain generalization is to obtain a model that achieves superior performance on testing data from unseen domains.  
\begin{figure*}
\centering
\subfloat[ERM]{
\includegraphics[width=0.24\textwidth]{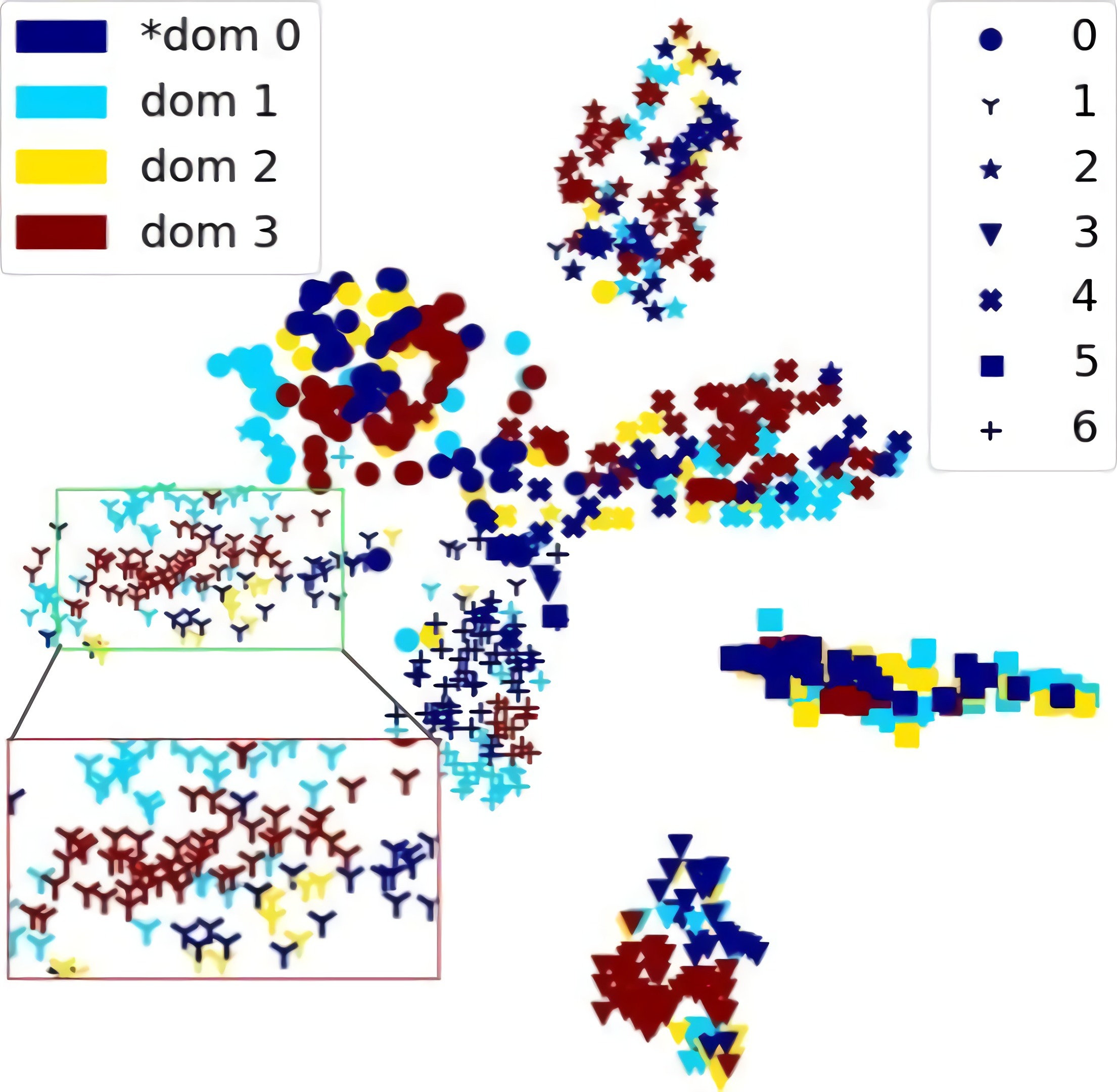}
}
\subfloat[DANN]{
\includegraphics[width=0.22\textwidth]{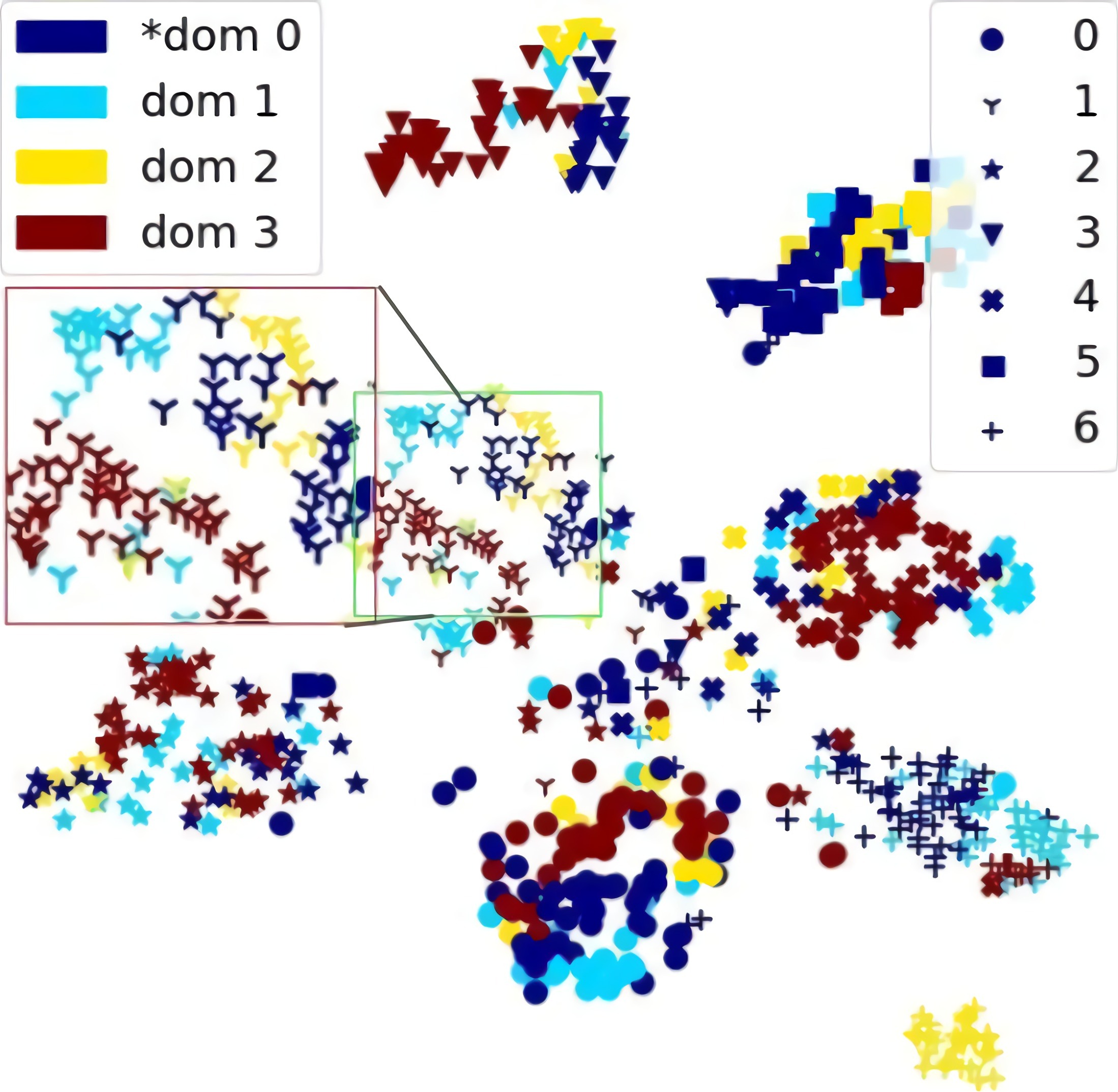}
}
\subfloat[CDANN]{
\includegraphics[width=0.22\textwidth]{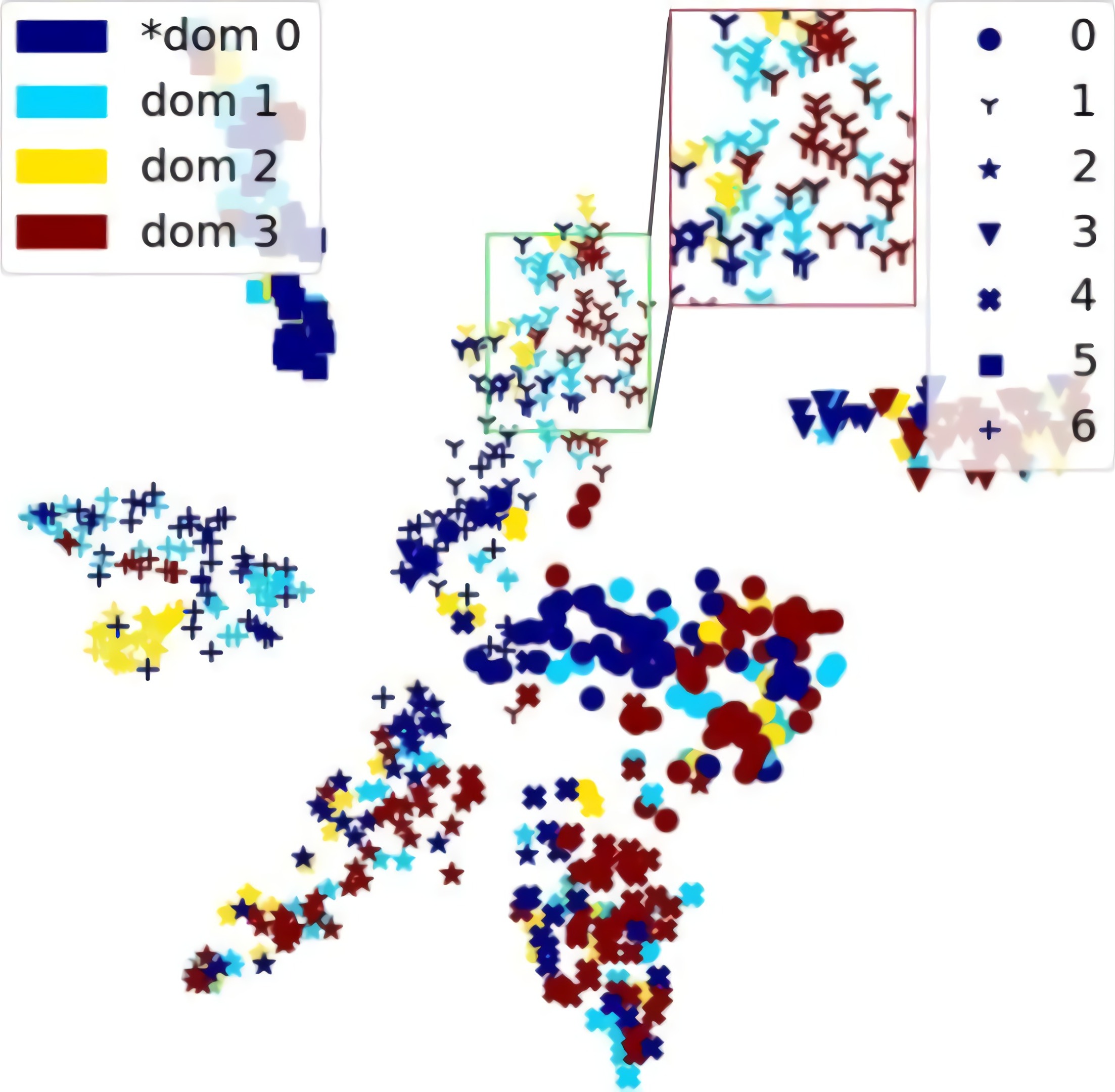}
}
\subfloat[Ours]{
\includegraphics[width=0.22\textwidth]{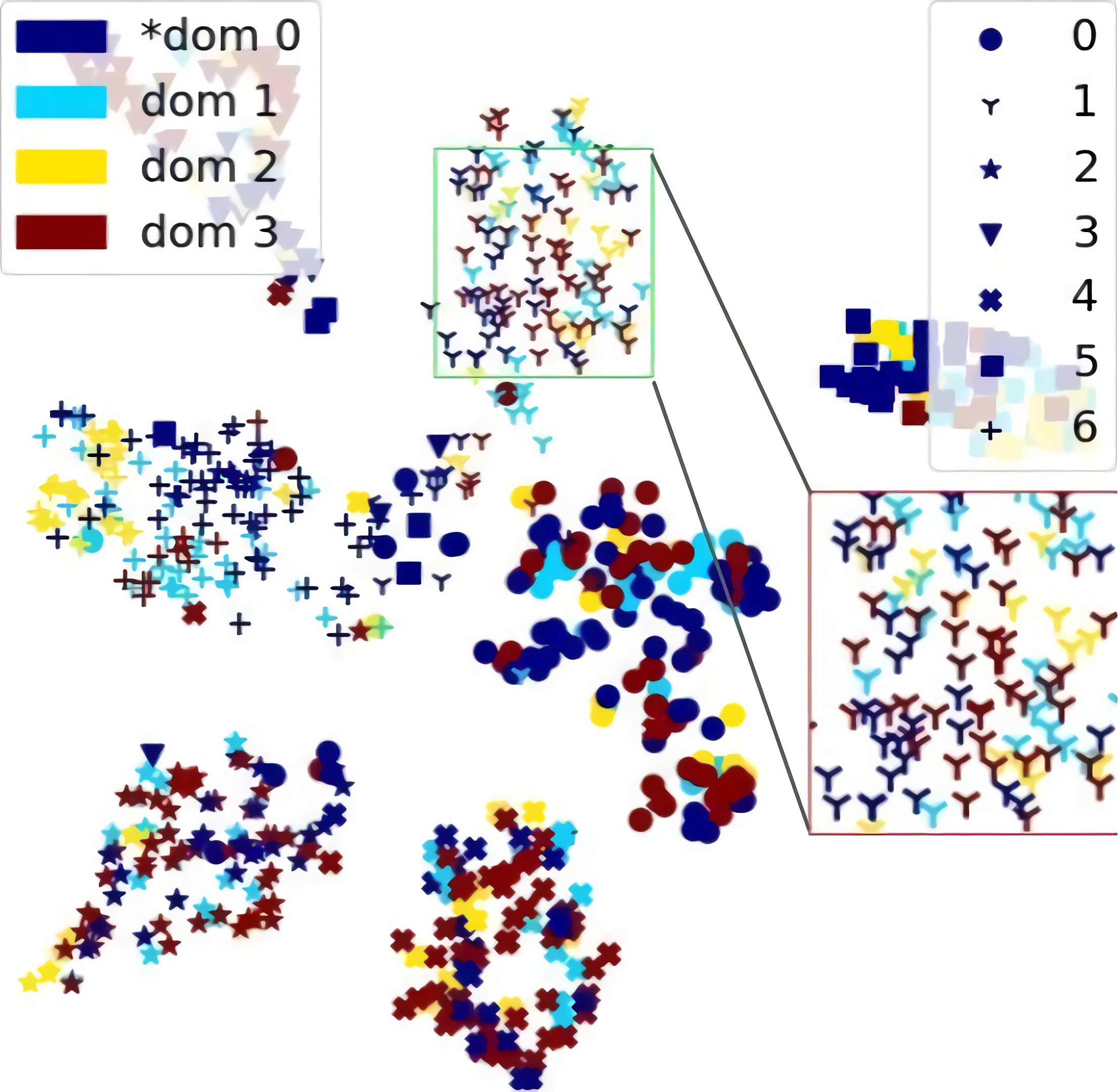}
}
\caption{Visualization of learned representation on PACS dataset. * denotes testing domain. Different shape(color) represents different class(domain).}
\label{fig:visEmbed}
\end{figure*}

\begin{figure*}
\centering
\subfloat[$V_{k}(H)$]{
\includegraphics[width=0.3\textwidth]{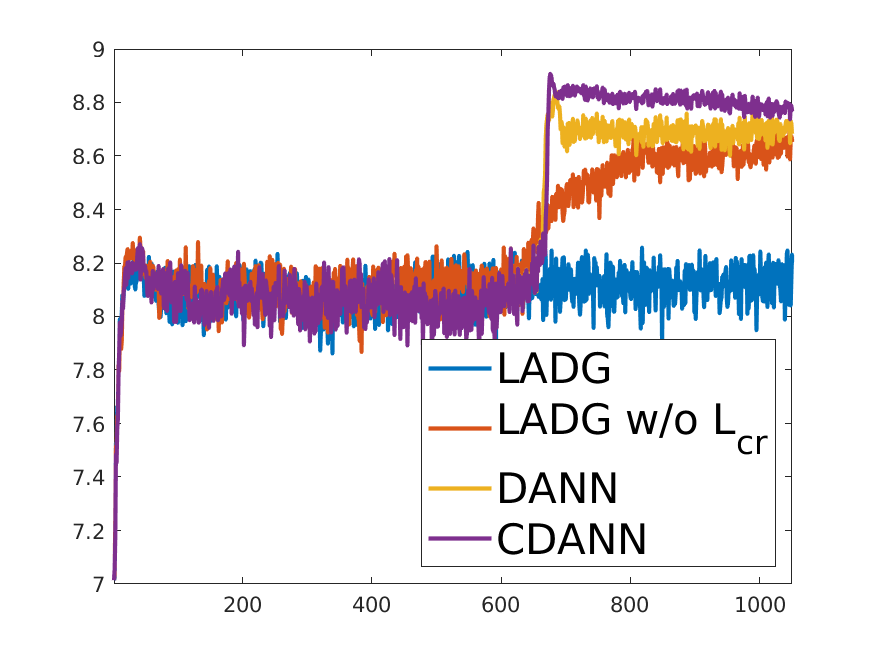}
}
\subfloat[$R(H)$]{
\includegraphics[width=0.3\textwidth]{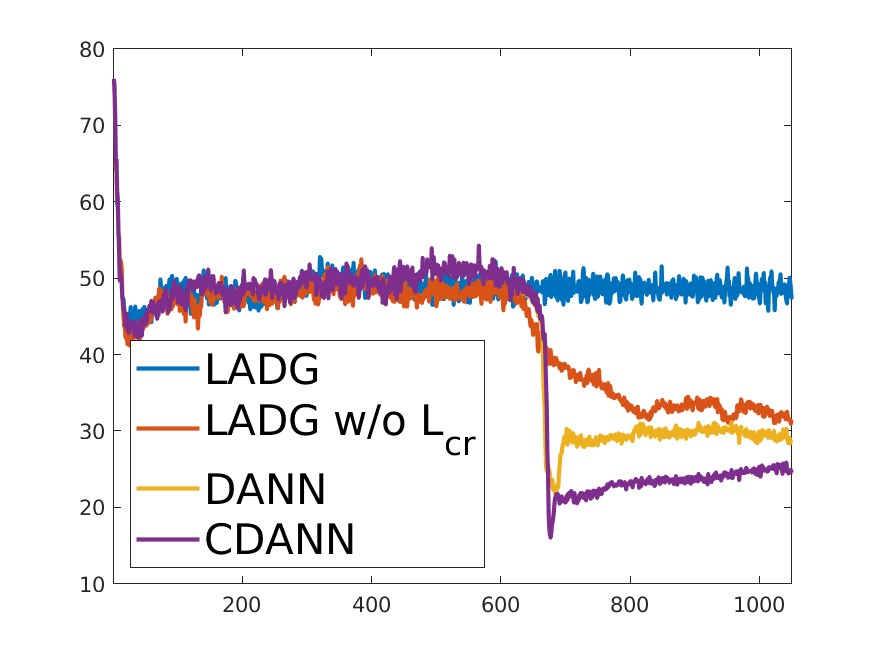}
}
\subfloat[$R_C(H)$]{
\includegraphics[width=0.3\textwidth]{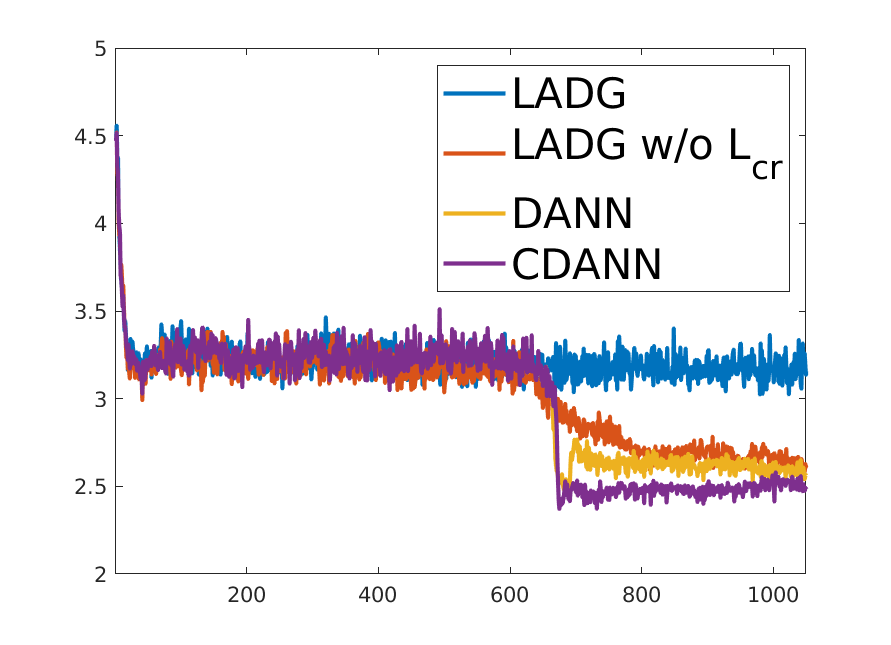}
}
\caption{The feature space will collapse with adversarial domain alignment. We pretrain the model for 600 steps with ERM and then ADG methods are applied. Our method trained with $\mathcal{L}_{cr}$ can alleviate space collapse.}
\label{fig:visCollapse}
\end{figure*}

A highly popular approach is adversarial domain generalization (ADG)~\cite{matsuura2020domain,sicilia2021domain,ganin2016domain,long2017conditional}. These are inspired by generative adversarial networks~\cite{goodfellow2014generative} and handle the problem by learning a domain-invariant representation with the aid of a domain discriminator, denoted as $\eta$. We use the seminal DANN~\cite{ganin2016domain} work as an exemplar to illustrate the approach. Given the training data, DANN trains the featurizer $\phi$ and the classifier $w$ to minimize the primary task loss:
\begin{equation} \label{eq:lt}
    \mathcal{L}_{t}(w, \phi) =  \frac{1}{n}\sum_{x_i,y_i \in \mathcal{D}_{tr}} \ell (w \circ \phi(x_i), y_i),
\end{equation}
where $\ell(\cdot)$ is the primary-task loss function. The featurizer representations likely overfit to the training domains and would struggle under the presence of domain shifts. This can lead to serious performance drops when testing the trained model on unseen domains. ADG methods aim to learn domain-invariant features to improve the generalization ability by using the domain discriminator, $\eta$, to distinguish samples from different domains using the extracted features $h_i$ by minimizing the domain classification loss:
\begin{equation}\label{eq:adv}
    \mathcal{L}_{dom}(\eta) = -\frac{1}{n}\sum_{x_i,e_i \in \mathcal{D}_{tr}} \log(p_i),
\end{equation}
where $p_i$ is the pseudo-probability for the ground-truth domain and the dependence on $x_{i}$ is implied. DANN uses the cross-entropy loss to train the domain classifier $\eta$. To obtain domain-invariant representations, the feature extractor $\phi$ is adversarially trained to maximize $L_{dom}$ so that the extracted features $H$ are indistinguishable to the domain classifier. We summarize the training objective as follows:
\begin{equation} \label{eq:dann}
\begin{split}
        \mathcal{L}_{gen} := \min_{w,\phi} ~ &\mathcal{L}_{t}(w, \phi) - \lambda \mathcal{L}_{dom}(\eta) \\
        \mathcal{L}_{disc} := \min_{\eta} ~ &\mathcal{L}_{dom}(\eta),     
\end{split}
\end{equation}
where $\lambda$ is used to balance the adversarial domain classification loss and the primary task loss.

\subsection{Limitations}
Although ADG methods have achieved progress, they do not show significant improvements over empirical risk minimization (ERM) according to recent benchmarks~\cite{gulrajani2020insearch}. We argue that two issues limit the performance of current ADG approaches: incomplete alignment and feature space over collapse. To illustrate this, we use three existing methods as exemplars: ERM, DANN~\cite{ganin2016domain}, and CDANN~\cite{long2017conditional}. DANN learns to match the feature distributions across domains $P(H^s))$ where $H^s=\phi(X^s)$, while CDANN aims to match the class-conditioned feature distributions. We provide observations and analysis using experiments conducted on the PACS dataset~\cite{li2017deeper}, which is a commonly used image classification dataset for domain generalization that comprises 7 different classes from four domains $S \in \lbrace art, cartoons, photo, sketches \rbrace$.


\subsubsection{Incomplete Alignment}

We train ADG methods under the benchmark setting with a \textit{linear predictor} $w$ and use tSNE~\cite{van2008visualizing} to visualize the learned representations. The results are shown in Fig. \ref{fig:visEmbed}. ADG methods operate under the assumption that the latent representation of samples from different training domains learned by ERM would be located at different sub-spaces, \ie{}, domain shift, and learning to align the data distributions across different domains would lead to a common representation space, \ie{}, a domain-invariant representation. However, observations from the visualization results of Fig. \ref{fig:visEmbed}a question this assumption. First, when using ERM, the features for different domains are surprisingly already roughly aligned and the domain shift is not as obvious as expected. Instead, samples from the same class are grouped together regardless of their domain labels. Second, the features from same domains often form localized clusters and the clusters will be spread over the whole space. The localized clusters are also cross-distributed across domains, which leads to localized domain-level statistical difference across domains. This suggests that ERM already incompletely aligns features, \ie{}, in a gross but non-local manner.


Since features are already at least partly aligned, the impact of additional measures that aim to align the domain-level distributions $P(H^s)$, \eg{}, DANN~\cite{ganin2016domain} and  MMD~\cite{li2018domain}, can be limited. In principle, ADG approaches could more completely align distributions across domains, \ie{}, produce local mixing without the localized clusters of Fig. \ref{fig:visEmbed}. Yet, in practice it is extremely difficult to train classification-based discriminators to learn a rich and complete enough representation of the domain distributions that actually describes the degree of local neighborhood-level mixing. As Fig. \ref{fig:visEmbed}b demonstrates, DANN features still exhibit the localized clusters of ERM. 


An intuitive way to alleviate this problem is to align the domain distributions conditioned on the class label $P(H^s_y)$ as adopted by CDANN~\cite{long2017conditional,li2018deep}. However, the conditioned domain distributions are still not well mixed (see Fig. \ref{fig:visEmbed}c) because of the intra-class intra-domain heterogeneity. Additionally, the class-conditioned domain alignment may partially dismiss the fact that the domain information contained in samples from different classes could mutually benefit the domain discrimination across classes. Finally, it is nontrivial to apply class-conditioned methods to general domain adaptation problems, \eg{}, regression or object detection.

To mitigate the problem, we introduce localized adversarial domain generalization (Section~\ref{sec:localadv}), which aligns the domain features at all local regions by incorporating a localized classifier with adversarial learning~\cite{bischl2013benchmarking}.

\subsubsection{Feature Space Collapse}
Another concern for ADG is feature space over collapse. An ideal result for ADG is to make features to be domain-invariant while maintaining the compactness of the feature space and keeping the primary task loss~$\mathcal{L}_t$ small. When updating the feature extractor $\phi$ to fool the domain discriminator by maximizing the domain classification loss $\mathcal{L}_{dom}$, a trivial way is to collapse the whole space or class-wisely collapse the feature space so that the $\mathcal{L}_{dom}$ could be maximized. To verify our assumption that the feature space tends to be over compact with ADG, we adopt three different strategies to measure the compactness of the learned feature space $H \in \mathcal{R}^{n \times d}$:
\begin{enumerate}
    \item We calculate the average degree of the k-nearest neighbors (KNN) graph. That is, we average the sum of cosine similarity between the $i$-th sample and its K-nearest neighbors:
    \begin{equation}
        V_{k}(H)=\frac{1}{N}\sum_{i=1}^N\sum_{j\in \mathcal{N}_i} \dfrac{\mathbf{h}_{i}^\tran \mathbf{h}_{j}}{\|\mathbf{h}_{i}\|\|\mathbf{h}_{j}\|},
    \end{equation} 
    where the set $\mathcal{N}_i$ represents the neighbors of the $i$-th sample. $S_{k}(H)$ can be used to monitor the local density;
    \item We calculate coding rate following~\cite{yu2020learning}:
    \begin{equation}
        R(H)=\frac{1}{2}\log \det(I+\frac{d}{n\epsilon^2}\mathbf{H}^\tran \mathbf{H}),
    \end{equation}
    where $\epsilon$ denotes the precision and we use $\mathbf{H}=[\mathbf{h}_1\ldots\mathbf{h}_n]^\tran$ to represent a set of features in matrix form. $R(H)$ indicates the number of bits needed to encode the data $H$ up to a precision $\epsilon$, and thus can be seen as a measurement of the compactness of the feature space. Please refer to ~\cite{yu2020learning,ma2007segmentation} for details. We normalize $H$ to eliminate scale effects when calculating $R(H)$;
    \item We average over the class-wise coding rate for the learned features as~\cite{yu2020learning}
    \begin{equation}
        R_C(H) =\sum_{y=1}^C \frac{N_y}{N} R(H^y, \epsilon).
    \end{equation}
    $R_C(H)$ can be used to measure the compactness for the class-wise feature space~\cite{yu2020learning}.
\end{enumerate}
For simplicity, we calculate all measurements with samples in the minibatch, and the results are shown in Fig. \ref{fig:visCollapse}. According to our experiments, when we train the model with ERM, all feature space compactness measurements remain stable. In contrast, we observe a sharp decrease of all measurements when we apply ADG. This implies that the features extracted by $\phi$ tend to collapse to trivially maximize the loss of the domain discriminator. The collapse often leads to overfitting and poses a threat to the generalization performance for unseen domains~\cite{yu2020learning}. While the feature space is expected to shrink with ADG by removing features correlated with domain information, the space is at high risk for over collapse. As shown in Fig. \ref{fig:visCollapse}c, we alleviate the over collapse by proposing a loss to encourage to maintain the compactness of the feature space.



\section{Our Method} 
\begin{figure}
\centering
\includegraphics[width=0.45\textwidth]{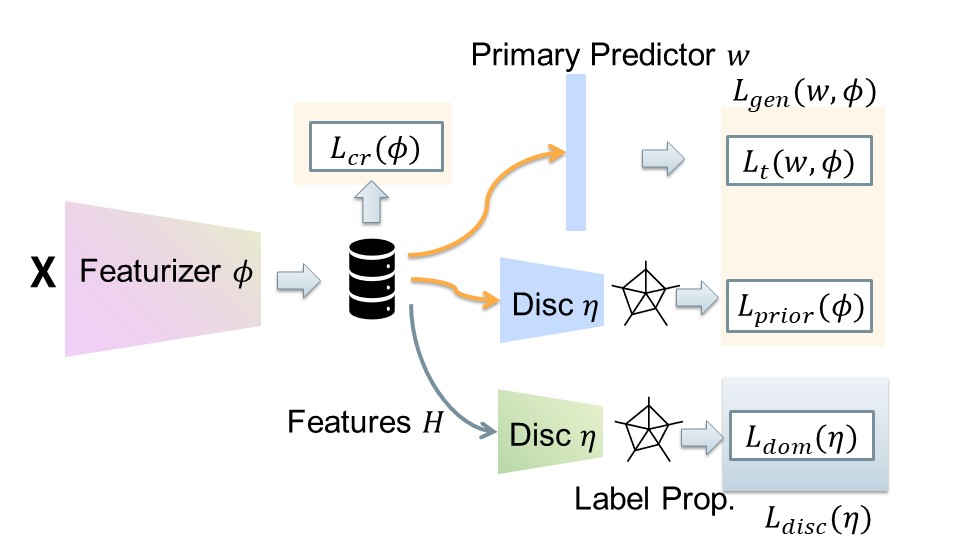}
\caption{Block diagram of our method.}
\label{fig:diag}
\end{figure}
We introduce localized adversarial domain generalization with space compactness maintenance~(LADG) to deal with the above two limitations. Fig. \ref{fig:diag} outlines our approach. LADG applies a localized classifier to examine every local region to achieve instance-wise domain alignment. Additionally, LADG adopts a coding rate inspired loss to prevent space collapse.

\subsection{Local Classification} \label{sec:localadv}

To deal with the problem of incomplete alignment, LADG adopts a local classifier to ADG. Local classifiers predict the label for each sample based on the local regions of interests around the target sample~\cite{bischl2013benchmarking}. The general idea here is that if a discriminator can correctly predict a sample's domain from its neighbors' domains, then local neighborhoods are not well mixed. Fooling such a discriminator can lead to more complete distributional alignment and better domain generalization performance. Typical local classification methods include KNN, decision trees, random forests~\cite{breiman2001random}, and localized logistic regression~\cite{hand2003local}. LADG instead adopts label propagation, which is effective, differentiable, and can be seamlessly incorporated with deep neural networks. Label propagation has been successfully applied to different problems, including few-shot learning~\cite{liu2018learning}, node classifiers~\cite{zhang2019adaptive}, metric learning~\cite{liu2019deep}, debiasing~\cite{zhu2021learning}, semi-supervised learning~\cite{iscen2019label}, and domain adaptation~\cite{cai2021theory,zhang2020label}, but is not well studied for domain generalization.


\begin{algorithm}[h]
\small
\DontPrintSemicolon
\SetAlgoLined
\SetNoFillComment
\LinesNotNumbered
 \KwData{the training set:$\mathcal{D}_{tr}$, $\rho=0.2$, $\xi=0.99$}
 \KwResult{the trained weight $\phi, w$}
 \While{the maximal iterations are not reached}{
    $\lbrace x_i, y_i, e_i\rbrace_{i=1}^{n} \sim \mathcal{D}_{tr}$ \tcp{sample minibatch}
    $\lbrace \mathbf{h}_i \rbrace_{i=1}^n = \lbrace \phi(x_i) \rbrace_{i=1}^n$ \tcp{extract features}
    \tcc{{Update Domain Disc $\eta$}}
    Obtain domain prediction $p_{ij}$ by Alg. \ref{alg:labelprop}\\
    Update $\eta$ by minimizing $\mathcal{L}_{disc}$ \eqref{eq:ladg}\\
    \tcc{Update $\phi$ and $w$}
    Obtain $p_{ij}$ by Alg. \ref{alg:labelprop} with updated $\eta$\\
    $\mathcal{L}_{t}(w, \phi) =  \frac{1}{n}\sum_{i=1}^n \ell (w \circ \phi(x_i), y_i)$~\eqref{eq:lt}\\
    $\mathcal{L}_{prior}(\phi) = -\frac{1}{n}\sum_{i=1}^n \sum_{j=1}^{S} q_j \log(p_{ij})$~\eqref{eq:maxentropy}\\
    $\mathcal{L}_{cr}(\phi) = \frac{1}{\rho}\log \cosh (\rho(R(H)-\bar{R}(H)))$~\eqref{eq:cr} \\
    Update $\phi$ and $w$ by minimizing $\mathcal{L}_{gen}$ \eqref{eq:ladg}\\
    $\bar{R}(H) \leftarrow \xi \bar{R}(H) + (1-\xi)R(H)$
 }

 \caption{Localized ADG}
 \label{alg:training}
\end{algorithm}

\begin{algorithm}[h]
\small
\DontPrintSemicolon
\SetAlgoLined
\SetNoFillComment
\LinesNotNumbered
 \KwData{Extracted features $\lbrace \mathbf{h}_i \rbrace_{i=1}^n$, domain label $\lbrace e_i\rbrace_{i=1}^{n}$}
 \KwResult{Domain prediction: $p_{ij}$}
$\mathbf{g}_i=\eta(\mathbf{h}_i)$ \tcp{Get domain features}
Construct similarity graph $\mathbf{A}$ by \eqref{eq:simgraph} \\  
$\mathbf{S}=\mathbf{D}^{-1/2} \mathbf{A} \mathbf{D}^{-1/2}$\\
Obtain converged prediction $\mathbf{R}^*$ by \eqref{eq:r}\\
$p_{ij} = \frac{exp(r_{ij}^*)}{\sum_{j=1}^S exp(r_{ij}^*)}$\\
 \caption{Label Propogation}
 \label{alg:labelprop}
\end{algorithm}

LADG trains a discriminator that adversarially projects feature representations into a new space, \ie{}, $\mathbf{g}_i = \eta(\mathbf{h}_i)$. We can use a standard MLP for the discriminator. Alternatively, other architectures are possible, \eg{}, GatedGCN~\cite{Bresson_2017} which allows the discriminator to exploit a sample's neighbors when projecting the features. This can give it additional information and a greater ability to discriminate, thereby further challenging the featurizer. Both options work well, with the latter providing a slight edge in our experiments. An affinity matrix, $\mathbf{A} = \lbrace a_{ij} \rbrace$, is then constructed based on the transformed features and their K-nearest neighbors: 
\begin{equation} \label{eq:simgraph}
a_{ij} = \begin{cases}
\exp \left(\dfrac{\tau\,\mathbf{g}_{i}^\tran \mathbf{g}_{j}}{2\|\mathbf{g}_{i}\|\|\mathbf{g}_{j}\|}\right) &\text{if}~j\in \mathcal{N}_i\\
0 &\text{otherwise}
\end{cases},
\end{equation}
where $\tau$ is a scale factor. Because it is intractable to construct an affinity matrix across an entire dataset, we instead construct one for each mini-batch. A Laplacian normalized similarity graph can then be obtained as $\mathbf{S}=\mathbf{D}^{-1/2} \mathbf{A} \mathbf{D}^{-1/2}$~\cite{chung1997spectral}, where the degree matrix $\mathbf{D}=\lbrace d_{ij} \rbrace$ is a diagonal matrix defined as $d_{ii} = \sum_j a_{ij}$.  We use the constructed graph to propagate the domain labels~\cite{zhang2019adaptive,liu2018learning,zhou2003learning}, here represented as an $n \times S$ matrix, $\mathbf{E}=\lbrace e_{ij} \rbrace$, where each row is a one-hot vector of the categorical domain label:
\begin{equation}
    \mathbf{R}^{t+1} = \alpha \mathbf{S} \mathbf{R}^{t} + (1-\alpha) \mathbf{E},
\end{equation}
where $\mathbf{R}^t$ are the propagated soft domain labels for the $t$-th propagation timestep, and $\alpha=0.8$ is the restart probability. We initialize $\mathbf{R}^0 \leftarrow \mathbf{E}$. The converged propagation results  can be calculated as~\cite{zhou2003learning}
\begin{equation} \label{eq:r}
    \mathbf{R}^* = (\mathbf{I}-\alpha \mathbf{S})^{-1}\mathbf{E}.
\end{equation}
We normalize the converged results into domain-prediction pseudo-probabilities: 
\begin{equation} \label{eq:pours}
    p_{ij} = \frac{\exp(r_{ij}^*)}{\sum_{j=1}^S \exp(r_{ij}^*)}.
\end{equation}
LADG updates the \emph{discriminator} by minimizing the domain classification loss $L_{dom}$ in \eqref{eq:adv} with the $p_{ij}$ defined in \eqref{eq:pours}. We conduct the label propagation with samples in the minibatch.
\begin{table}[]
\centering
\begin{tabular}{@{}lcccc@{}}
\toprule
dataset       & task & input & \# samples & \# domains \\ \midrule
iWildCam      & Cls. & image & 203,029    & 323        \\
Camelyon17    & Cls. & image & 455,954    & 5          \\
PovertyMap    & Reg. & image & 19,669     & 23x2       \\
FMoW          & Cls. & image & 523,846    & 16x5       \\
CivilComments & Cls. & text  & 448,000    & 16         \\
Amazon        & Cls. & text  & 539,502    & 2586       \\ \bottomrule
\end{tabular}
\caption{Wilds datasets statistics.}
\label{tab:datawilds}
\end{table}

Analogously, we could follow standard ADG practices and simply train the featurizer to produce domain-invariant representations by maximizing the same $L_{dom}$ in \eqref{eq:dann}. However, we found that this will not only make the training process unstable, but also likely make the domains trivially aligned when the number of training domains $S\geq 3$. To see this, assume four domains as $\lbrace A, B, C, D \rbrace$. We could trivially update the featurizer $\phi$ to fool the discriminator by aligning $A$ with $B$, and $C$ with $D$, without even aligning the $A$ and $C$ domains together. Instead we need a loss that directly formulates our end goal of local neighborhood mixing. To do this, we calculate a prior distribution of domain labels for each minibatch, $q_j = \frac{1}{n} \sum_{i=1}^N e_{ij}$. The featurizer is then trained to match each sample's propagated domain pseudo-probability with the prior distribution: 
\begin{equation} \label{eq:maxentropy}
   \mathcal{L}_{prior}(\phi) = -\frac{1}{n}\sum_{i=1}^n \sum_{j=1}^{S} q_j \log(p_{ij}),
\end{equation}
thus encouraging each local neighborhood to be well mixed with all domains. We empirically validate the effectiveness of \eqref{eq:maxentropy} in the experiments. 

\subsection{Alleviate Space Over Collapse}\label{sec:space}
The featurizer may trivially minimize the feature space to make the domains less distinguishable, \ie{}, space collapse, which will dramatically degrade the generalization performance for testing domains. To alleviate this problem, we penalize feature space over collapse by encouraging to maintain the compactness for the feature space $R(H)$:
\begin{equation} \label{eq:cr}
\mathcal{L}_{cr}(\phi) = \frac{1}{\rho}\log \cosh (\rho(R(H)-\bar{R}(H))).
\end{equation}
We elaborate on several points for \eqref{eq:cr}. First, we calculate the coding rate $R(H)$ within each minibatch for tractability. Second, $\bar{R}(H)$ is the moving average of $R(H)$ across batches, and is updated as $\bar{R}(H) \leftarrow \xi \bar{R}(H) + (1-\xi)R(H)$, where $\xi=0.99$. $\bar{R}(H)$ can be initialized during pretraining. We adopt a Log-Cosh loss instead of $L_1$, which makes the loss effectively zero beyond a certain tolerance value, allowing the network to focus on more egregious violations. A hyperparameter $\rho$ further smoothes the loss curve, which we set to $0.2$. Lastly, we use the overall coding rate $R(H)$ instead of the class-wise one $R_y(H)$ since we observe that stable $R(H)$ directly leads to stable $R_y(H)$ and it is nontrivial to calculate the class-wise coding rate $R_y(H)$ for general primary tasks. 

\subsection{Overall Training}
We summarize the overall training objective of LADG as 
\begin{equation} \label{eq:ladg}
\begin{split}
        \mathcal{L}_{gen} := \min_{w,\phi} ~ &\mathcal{L}_{t}(w, \phi) + \lambda \mathcal{L}_{prior}(\phi) + \gamma \mathcal{L}_{cr}(\phi)\\
        \mathcal{L}_{dom} := \min_{\eta} ~ &\mathcal{L}_{adv}(\eta),     
\end{split}
\end{equation}
where $\lambda$ and $\gamma$ are used to balance different terms. As the domain discriminator will  be  discarded  during inference, our method introduces no extra parameters and computational cost compared with ERM for testing. We summarize the training process of LADG in Algorithm \ref{alg:training}. Because we calculate affinities across mini-batches, the sampling strategy plays an important role. We randomly select $K$ domains and populate the mini-batch with an equal number of samples from each domain.  

\begin{table*}[]
\centering
\begin{tabular}{@{}lcccccc@{}}
\toprule
                          & iWildCam  & Camelyon17 & PovertyMap  & FMoW       & CivilComments & Amazon     \\  
\multirow{-2}{*}{Methods} & F1        & avg acc    & wg r        & wg acc     & wg acc        & 10\% acc   \\ \midrule
ERM                       & 31.0(1.3) & 70.3(6.4)  & 0.45(0.06)  & 32.3(1.25) & 56.0(3.6)     & \textbf{53.8}(0.8)  \\
CORAL                     & \underline{32.8(0.1)} & 59.5(7.7)  & 0.44(0.06)  & 31.7(1.24) & 65.6(1.3)     & 52.9(0.8)  \\
IRM                       & 15.1(4.9) & 64.2(8.1)  & 0.43 (0.07) & 30.0(1.37) & 66.3 (2.1)    & 52.4 (0.8) \\
Group DRO & 23.9 (2.1) & 68.4 (7.3) & 0.39 (0.06) & 30.8 (0.81) & \underline{70.0}(2.0) & \underline{53.3 (0.0)} \\
FISH      & 22.0 (1.8) & \underline{74.7 (7.1)} & 0.30 (0.01)  & \textbf{34.6} (0.18) & \textbf{73.1}* (1.2) & \underline{53.3* (0.0)} \\  \hline
DANN                      & 27.9(1.1) & 65.6(6.7)  & \underline{0.45(0.06)}  & 30.6(0.15) &    66.0(1.0)           & \underline{53.3(0.0)}           \\
CDANN                     &  \underline{32.7(1.5)}         & 66.3(6.1)  & n/a            & 33.1(0.11) & 69.5(1.5)              &  \underline{53.3(0.0)}          \\
LADG                      & \textbf{33.1}(0.6) & \textbf{76.5}(7.7)  & \textbf{0.48}(0.07)  & \underline{{33.5}(2.08)} &  66.9(1.3)             & \underline{53.3(0.0)}           \\ \bottomrule
\end{tabular}%
\caption{Out-of-distribution test results on Wilds. wg means worst group performance.  * FISH adopts different backbone for CivilComments and Amazon~\cite{fishshi2021gradient}. Best results are highlighted. DANN and CDANN are implemented by ourselves.}
\label{tab:all}
\end{table*}

\begin{table}[]
\centering
\begin{tabular}{@{}lcc@{}}
\toprule
Camelyon17  & val acc    & test acc*    \\ \midrule
ERM       & 84.9(3.1)  & 70.3(6.4)   \\
CORAL     & \textbf{86.2}(1.4)  & 59.5(7.7)   \\
IRM       & \textbf{86.2}(1.4)  & 64.2(8.1)   \\
Group DRO & 85.5(2.2)  & 68.4(7.3)   \\
FISH      & 83.9(1.2)  & \underline{74.7(7.1)}   \\ \hline
DANN      & 85.6(1.1)     &65.6(6.7) \\
CDANN      & \underline{86.0(1.3)}     &66.3(6.1) \\
LADG      & {85.4}(0.8) & \textbf{76.5}(7.7) \\ \bottomrule
\end{tabular}%
\caption{Results on Camelyon17. Validation results are OOD. Test acc is the main summary metric.}
\label{tab:came}
\end{table}

\begin{table}[]
\centering
\begin{tabular}{@{}lcccc@{}}
\toprule
iWildCam  & ID F1 & ID acc & OOD F1* & OOD acc \\ \midrule
ERM       & \textbf{47.0}(1.4)        & \textbf{75.7}(0.3)       & 31.0(1.3)         & 71.6(2.5)        \\
CORAL     & 43.5(3.5)        & 73.7(0.4)       & \underline{32.8(0.1)}         & \underline{73.3(4.3)}        \\
IRM       & 22.4(7.7)        & 59.9(8.1)       & 15.1(4.9)         & 59.8(3.7)        \\
Group DRO & 37.5(1.7)        & 71.6(2.7)       & 23.9(2.1)         & 72.7(2.0)        \\
FISH      & 40.3(0.6)        & \underline{73.8(0.1)}       & 22.0(1.8)         & 64.7(2.6)        \\ \hline
DANN      & 42.3(0.5)        & 73.5(0.1)       & 27.9(1.1)         & 71.8(1.3)        \\
CDANN      & \underline{46.7(0.7)}        & 75.0(0.2)       & 32.7(1.5)         & 72.9(1.5)        \\
LADG      & {46.4}(1.2)        & {73.7(0.8)}       & \textbf{33.1}(0.6)         & \textbf{74.4}(2.7)        \\ \bottomrule
\end{tabular}%
\caption{Results on iWildCam. All metrics are from the test set. OOD F1 is the main summary metric.}
\label{tab:iwildcam}
\end{table}

\section{Experiments}
As has been noted, robustness to synthetic distributional shifts do not necessarily translate to robustness to real-world shifts~\cite{taori_measuring_2020, djolonga_robustness_2021}. For this reason, we conduct experiments primarily on the Wilds benchmark~\cite{koh2021wilds}, which is composed of real-world datasets representing natural distributional shifts from many different modalities and applications, \eg{}, biomedical imagery, satellite imagery, and text. Importantly, all Wilds datasets reflect distributional shifts that produce significant performance drops between in-distribution (ID) and out-of-distribution (OOD) data~\cite{koh2021wilds}. We follow the experimental settings of FISH~\cite{fishshi2021gradient}, a recent and leading approach, and test on the Poverty~\cite{yeh2020using}, Camelyon17~\cite{bandi2018detection}, FMoW~\cite{christie2018functional}, CivilComments~\cite{borkan2019nuanced}, iWildCam~\cite{beery2021iwildcam}, and Amazon~\cite{ni2019justifying} Wilds datasets. \emph{Note CivilComments is designed for sub-population shifts (training and testing domains overlap), so it is not a DG dataset, but we include it for completeness.}  We summarize the statistics of the used datasets in Table~\ref{tab:datawilds}, and we refer the readers to \cite{koh2021wilds} for details on these datasets. We report the Wilds leaderboard metrics, which always include OOD test, but also may include ID test or OOD validation metrics.


\begin{table}[]
\centering
\resizebox{\columnwidth}{!}{
\begin{tabular}{@{}lcccc@{}}
\toprule
PovertyMap      & val r      & test r     & val wg r   & test wg r*  \\ \midrule
ERM          & 0.80(0.04)  & 0.78(0.04) & 0.51(0.06) & 0.45(0.06) \\
CORAL        & 0.80(0.04)  & 0.78(0.05) & 0.52(0.06) & 0.44(0.06) \\
IRM          & \underline{0.81(0.03)} & 0.77(0.05) & \underline{0.53(0.05)} & 0.43(0.07) \\
Group DRO    & 0.78(0.05) & 0.75(0.07) & 0.46(0.04) & 0.39(0.06) \\
FISH         & \textbf{0.82}(0.0)  & \textbf{0.80(0.02)}  & 0.47(0.01) & 0.30(0.01)  \\ \hline
DANN  & 0.80(0.04) & \underline{0.79(0.03)} & 0.52(0.05) & \underline{0.45(0.06)} \\
CDANN  & n/a & n/a& n/a & n/a\\
LADG & 0.80(0.03) & \underline{0.79(0.04)} & {0.51}(0.07) & \textbf{0.48}(0.07) \\ \bottomrule
\end{tabular}%
}
\caption{Results on PovertyMap. CDANN is not applicable for regression. r denotes Pearson correlation coefficient. All validation metrics are OOD. wg means worst-group performance and test wg r coefficient is the main summary metric.}
\label{tab:poverty}
\end{table}

\begin{table}[]
\centering
\resizebox{0.5\textwidth}{!}{
\begin{tabular}{@{}lcccc@{}}
\toprule
FMoW      & val acc     & test acc    & val wr acc   & test wr acc* \\ \midrule
ERM       & \textbf{59.5(0.37)}  & \textbf{53.0(0.55)}  & 48.9(0.62)   & 32.3(1.25)  \\
CORAL     & 56.9(0.25)  & 50.5(0.36)  & 47.1(0.43)   & 31.7(1.24)  \\
IRM       & 57.4(0.37)  & 50.8(0.13)  & 47.5(1.57)   & 30.0(1.37)  \\
Group DRO & \underline{58.8}(0.19)  & \underline{52.1}(0.5)   & 46.5(0.25)   & 30.8(0.81)  \\
FISH      & 57.8(0.15)  & 51.8(0.32)  & 49.5(2.34)   & \textbf{34.6}(0.18)  \\ \hline
DANN      & 57.1(0.73)  & 50.2(0.36)  & \underline{50.2(1.01)}   & {30.6}(0.15)  \\
CDANN      & 56.8(0.11)  & 51.0(0.36)  & 50.0(0.31)   & {33.1}(0.11)  \\
LADG      & 57.2(0.15) & 51.2(0.24) & \textbf{51.0} (0.47) & \underline{33.5(2.08)} \\ \bottomrule
\end{tabular}%
}
\caption{Results on FMoW. All validation results are OOD. wr means worst-region performance and test wr acc is the main summary metric.}
\label{tab:fmow}
\end{table}

\begin{table}[]
\centering
\resizebox{\columnwidth}{!}{
\begin{tabular}{@{}lcccc@{}}
\toprule
Civ.Comm. & val avg acc & val wg acc & test avg acc & test wg acc* \\ \midrule
ERM           & \textbf{92.3}(0.2)   & 50.5(1.9)  & \textbf{92.2}(0.1)    & 56.0(3.6)   \\
CORAL         & 88.9(0.6)   & 64.7(1.4)  & 88.7(0.5)    & 65.6(1.3)   \\ 
IRM           & 89.0(0.7)   & 65.9(2.8)  & 88.8(0.7)    & 66.3(2.1)   \\
Group DRO     & \underline{90.1}(0.4)   & 67.7(1.8)  & \underline{89.9}(0.5)    & \underline{70.0}(2.0)   \\
FISH*          & 89.7*(0.3)   & \textbf{72.0}*(1.0)  & 89.5*(0.2)    & \textbf{73.1}*(1.2)   \\ \hline
DANN          & 89.1(0.5)   & 68.8(1.1)  & 88.9(0.5)    & 66.0(1.0)   \\
CDANN          & 89.6(0.2)   & \underline{69.4}(0.5)  & 89.3(0.2)    & 69.5(1.5)   \\
LADG          & 88.8(0.6)   & 66.8(2.1)  & 88.6(0.6)    & 66.9(1.3)   \\ \bottomrule
\end{tabular}%
}
\caption{Results on CivilComments. * FISH uses a different backbone~\cite{fishshi2021gradient}. For CivilComments, the training and test domains overlap and there is only sub-population shift. wg means worst-group performance and test wg acc is the main summary metric.}
\label{tab:CivilComments}
\end{table}

\begin{table}[]
\centering
\resizebox{\columnwidth}{!}{
\begin{tabular}{@{}lcccc@{}}
\toprule
Amazon    & val acc   & test acc  & val 10\% acc & test 10\% acc \\ \midrule
ERM       & \textbf{72.7}(0.1) & \textbf{71.9}(0.1) &\textbf{ 55.2}(0.7)  & \textbf{53.8}(0.8)   \\
CORAL     & 72.0(0.3) & 71.1(0.3) & \underline{54.7(0.1)}  & \underline{53.3(0.1)}   \\
IRM       & 71.5(0.3) & 70.5(0.3) & 54.2(0.8)  & 52.4(0.8)   \\
Group DRO & 70.7(0.6) & 70.0(0.6) & \underline{54.7(0.0)}  & \underline{53.3(0.0)}   \\
FISH*      & \underline{72.5*(0.0)} & \underline{71.7*(0.1)} & \underline{54.7*(0.0)}  & \underline{53.3*(0.0)}   \\ \hline
DANN      & 72.1(0.0)     &71.4(0.1)     & \underline{54.7(0.0)}      & \underline{53.3(0.0)}       \\
CDANN     & 72.3(0.2)     & 71.2(0.2)     & \underline{54.7(0.0)}     & \underline{53.3(0.0)}       \\
LADG      & 71.9(0.1)     & 71.1(0.2)     & \underline{54.7(0.0)}      & \underline{53.3(0.0)}       \\ \bottomrule
\end{tabular}%
}
\caption{Results on Amazon. * FISH uses a different backbone~\cite{fishshi2021gradient}. Validaton metrics are OOD and test 10\% acc is the main summary metric.}
\label{tab:amazon}
\end{table}

\noindent{\textbf{Experimental Settings:}}
We follow the settings and evaluation metrics of Wilds benchmark~\cite{koh2021wilds}. We compare our methods with ERM~\cite{koh2021wilds}, Coral~\cite{sun2017correlation}, Group DRO~\cite{sagawa2019distributionally}, IRM~\cite{arjovsky2019invariant}, and Fish~\cite{fishshi2021gradient}, which we retrieve from the Wilds benchmark leaderboard~\cite{koh2021wilds}. We implement DANN~\cite{ganin2016domain} and CDANN~\cite{long2017conditional} ourselves. We adopt the same featurizer $\phi$, predictor $w$, and model selection strategies used in the Wilds benchmark. Following the Wilds specifications~\cite{koh2021wilds}, for iWildCam, FMoW, and Amazon, we average the results over three different runs, and for Camelyon17~(CivilComment), we run the experiments ten~(five) times. For PovertyMap, we average the results over five different folds.






\noindent{\textbf{Results:}} We summarize the results in Table~\ref{tab:all} for the main summary metrics, and detailed results for each dataset are provided in Tables~\ref{tab:came}--\ref{tab:amazon}. From Table~\ref{tab:all}, LADG  achieves state-of-the art results on 3 datasets and the second-best result on FMoW. Notably, LADG performance is much more stable, with the relative (and rank-order) performance of some methods, \eg{} FISH, varying considerably from dataset to dataset. Despite not being designed to address sub-population shifts, LADG still exhibits good performance on CivilComments, outperforming ERM. All methods perform comparably for Amazon (likely because its huge numbers of domains make it difficult to beat ERM), but LADG matches the best DG methods' performances. These results demonstrate that for true DG dataset, LADG (1) exhibits leading performance; and (2) its performance is consistently high. 

\noindent{\textbf{Ablation Study:}}
In  this  section,  we  conduct  ablation studies on Camelyon17 and Poverty Map to validate the different components of the proposed methods. The experiments follow the exact same settings as Wilds benchmark. The results are shown in Table \ref{tab:ablations}. DANN w/ $\mathcal{L}_{cr}$ equips DANN with $\mathcal{L}_{cr}$ loss. LADG w/o $\mathcal{L}_{prior}$ replaces prior matching loss with $\mathcal{L}_{adv}$ and LADG w/o GNN replace GNN with MLP. The experiments show that the coding ratio loss $\mathcal{L}_{cr}$ is important for our method, and LADG outperforms its direct counterpart DANN either with or without $\mathcal{L}_{cr}$ loss.  
\begin{table}[]
\centering
\resizebox{0.4\textwidth}{!}{%
\begin{tabular}{@{}lccc@{}}
\toprule
            & Camelyon17 & \multicolumn{2}{c}{PovertyMap} \\ 
            & avg acc    & r              & wg r          \\ \midrule
DANN  & 65.6(6.7)  &0.79(0.03)      &0.45(0.06)     \\ 
DANN w/ $\ell_{cr}$ & 73.8(7.9)  &0.78(0.04)      &0.47(0.06)     \\ \hline
LADG w/o $\mathcal{L}_{cr}$    & 68.8(5.4)  & 0.79(0.05)     & 0.43(0.07)    \\
LADG w/o $\mathcal{L}_{prior}$ & 72.8(7.3)  & \textbf{0.80}(0.07)     & 0.46(0.13)    \\
LADG w/o GNN     & 75.7(6.3)  & 0.79(0.05)     & \textbf{0.49(0.03)}    \\ 
LADG        & \textbf{76.5(7.7)}  & {0.79(0.04)}     & 0.48(0.07)    \\
\bottomrule
\end{tabular}%
}
\caption{Ablation studies on Wilds.}
\label{tab:ablations}
\end{table}
\vspace{-2mm}

\section{Limitations and Future Work}

Our analysis and experiments are conducted under the benchmark settings with a linear predictor $w$~\cite{gulrajani2020insearch,koh2021wilds}, and we do not experimentally verify the performance of our method under the settings with non-linear predictors. We expect incompletely aligned distributions with unmixed local neighborhoods to also be a problem when using non-linear predictors and that LADG's innovations would equally apply. We leave this for future work. Another limitation is that our method treats each domain as equally important. In practice, we may over-sample the domains that are deviated from others to improve the worst group performance for better generalization ability~\cite{sagawa2019distributionally}. Finally, improving performance when domains numbers are huge is another important avenue for future work.  

\section{Conclusions}
We proposed LADG, which addresses two limitations of existing ADG approaches: (1) incomplete distributional alignment with unmixed neighorhoods and (2) feature space over collapse. LADG incorporates an adversarial localized classifier to align domains within every local region. Moreover, LADG penalizes space over collapse by encouraging to maintain the compactness of the feature space via a differentiable coding-rate formulation. Experiments on leading benchmarks validate LADG's effectiveness and consistent performance across datasets.    

{\small
\bibliographystyle{ieee_fullname}
\bibliography{egbib}
}

\end{document}